\documentclass{article} % For LaTeX2e
\usepackage{iclr2026_conference,times}

\usepackage{amsmath,amsfonts,bm}

\def\eqref#1{equation~\ref{#1}}
\def\1{\bm{1}}

\DeclareMathAlphabet{\mathsfit}{\encodingdefault}{\sfdefault}{m}{sl}
\SetMathAlphabet{\mathsfit}{bold}{\encodingdefault}{\sfdefault}{bx}{n}

\usepackage{hyperref}
\usepackage{url}
\usepackage{booktabs} 
\usepackage{multirow} 
\usepackage{makecell}
\usepackage{graphicx}
\usepackage{amsmath}
\usepackage{amsfonts}       % blackboard math symbols
\usepackage{nicefrac}       % compact symbols for 1/2, etc.
\usepackage{microtype}      % microtypography
\usepackage{xcolor}         % colors
\usepackage{arydshln}
\usepackage{float}
\usepackage{pifont}
\usepackage[ruled,vlined]{algorithm2e}
\usepackage{lipsum} % just for position
\usepackage{amsmath}
\usepackage{xcolor}
\usepackage{wrapfig}
\usepackage[most]{tcolorbox}
\usepackage{enumitem}
\usepackage{titlesec}
\usepackage{titletoc}
\usepackage[table]{xcolor}

\usepackage{fontawesome}
\newcommand{\dataset}{\textsc{ReasonMap}\xspace}
\newcommand{\datasetplus}{\textsc{ReasonMap-Plus}\xspace}
\newcommand{\method}{\textsc{RewardMap}\xspace}
\newcommand{\revise}[1]{\textcolor{black}{#1}}

\definecolor{map_red}{RGB}{239,99,75}
\definecolor{map_blue}{RGB}{99,113,250}
\definecolor{map_green}{RGB}{0,180,139}
\definecolor{map_yellow}{RGB}{229,157,35}
\definecolor{map_gray}{RGB}{165,165,165}
\definecolor{link}{RGB}{229,158,221}
\definecolor{mygreen}{RGB}{93,173,85}
\definecolor{myred}{RGB}{192,57,43}
\newcommand{\resup}[2]{%
  #1 {\fontsize{7.5pt}{1em}\selectfont\textcolor{mygreen}{$\!\uparrow\!$ \textbf{#2}}}%
}

\hypersetup{
  colorlinks=true,
  linkcolor={red!70!black},
  citecolor=gray,
  urlcolor=cyan,
}

\title{\method: Tackling Sparse Rewards in Fine-grained Visual Reasoning via Multi-Stage Reinforcement Learning}
\author{Sicheng Feng$^{1,\dagger}$,~ 
Kaiwen Tuo$^{1,2,\dagger}$,~ 
Song Wang$^3$,~ 
Lingdong Kong$^4$,~
Jianke Zhu$^3$, 
\\
\textbf{Huan Wang}$^{1,*}$ 
\\
$^1$Westlake University \quad $^2$Tongji University \quad $^3$Zhejiang University 
\\
$^4$National University of Singapore
\\[0.5ex]
$^\dagger${\small Equal contributions.} \quad $^*${\small Corresponding author.}
\\[1ex]
\faGithubAlt~\textbf{Dataset \& Toolkit:} \href{https://fscdc.github.io/RewardMap}{\texttt{https://fscdc.github.io/RewardMap}}
}

\iclrfinalcopy % Uncomment for camera-ready version, but NOT for submission.
\begin{document}

\maketitle

\begin{abstract}

Fine-grained visual reasoning remains a core challenge for multimodal large language models (MLLMs). The recently introduced \dataset highlights this gap by showing that even advanced MLLMs struggle with spatial reasoning in structured and information-rich settings such as transit maps, a task of clear practical and scientific importance.
However, standard reinforcement learning (RL) on such tasks is impeded by sparse rewards and unstable optimization.
To address this, we first construct \datasetplus, an extended dataset that introduces dense reward signals through Visual Question Answering (VQA) tasks, enabling effective cold-start training of fine-grained visual understanding skills.
Next, we propose \method, a multi-stage RL framework designed to improve both visual understanding and reasoning capabilities of MLLMs.
\method incorporates two key designs. 
\textbf{First}, we introduce a difficulty-aware reward design that incorporates detail rewards, directly tackling the sparse rewards while providing richer supervision. 
\textbf{Second}, we propose a multi-stage RL scheme that bootstraps training from simple perception to complex reasoning tasks, offering a more effective cold-start strategy than conventional Supervised Fine-Tuning (SFT).
Experiments on \dataset and \datasetplus demonstrate that each component of \method contributes to consistent performance gains, while their combination yields the best results.
Moreover, models trained with \method achieve an average improvement of $3.47\%$ across $6$ benchmarks spanning spatial reasoning, fine-grained visual reasoning, and general tasks beyond transit maps, underscoring enhanced visual understanding and reasoning capabilities.

\end{abstract}
\section{Introduction}
\label{sec:intro}

Fine-grained visual reasoning over structured visual inputs remains a significant challenge for multimodal large language models (MLLMs)~\citep{bai2025qwen25,openai2025o3,zhang2025thyme,worldlens,li2025_3eed,li2025seeground}. 
Recently, \dataset~\citep{feng2025reasonmap} was introduced as a benchmark on high-resolution transit maps in the real world, where tasks (\textit{e.g.}, route planning) combine visual understanding with spatial reasoning, jointly constituting the fine-grained visual reasoning challenge \citep{xie2025drivebench,survey_vla4ad,survey_3d_4d_world_models,kong2025talk2event}.
This task is not only of practical value for real-world navigation and transportation systems, but also of fundamental scientific interest as it exposes reasoning gaps in current MLLMs.
Despite steady advances in vision–language pre-training~\citep{liu2023llava,liu2024llavanext,bai2025qwen25}, existing models consistently struggle with the visual and spatial reasoning demands in \dataset. 
This gap motivates us to investigate how reinforcement learning (RL)~\citep{zhang2025surveyrl, shao2024deepseekmath, guo2025deepseekr1} can be adapted to enhance fine-grained visual reasoning abilities in structured visual domains such as transit maps.

However, directly applying standard RL methods~\citep{rafailov2023direct, shao2024deepseekmath} to complex tasks such as \dataset is highly challenging, as supervision signals are inherently sparse, \textit{i.e.}, success is typically judged only at the final answer after a long reasoning chain~\citep{quadros2025llm, cao2024beyond, chen2025reasoning}. The difficulty of the tasks further amplifies this sparsity, which in turn destabilizes optimization and hinders effective exploration~\citep{wang2025harnessing}. 
While classical approaches like Supervised Fine-Tuning (SFT)~\citep{liu2023llava,wei2025advancing} offer dense supervision, they fall short in equipping models for the long-chain decision-making intrinsic to visual reasoning tasks.
This mismatch between task complexity and supervision signal forms a critical bottleneck in leveraging RL for fine-grained visual reasoning.

To address this issue, we first construct \datasetplus, an extended dataset that introduces dense reward signals for further cold-start training. Tasks in \datasetplus are organized along a natural difficulty continuum, from simple Visual Question Answering (VQA) that enhances perception to progressively harder visual tasks reflecting the complexity of fine-grained visual reasoning queries. 
We further introduce \method, a multi-stage RL framework with detailed reward design. It consists of two key components: (1) a reward scheme that, beyond basic format and correctness rewards, incorporates detail rewards to mitigate sparsity in supervision for hard samples and adopts a difficulty-aware design to account for task complexity; and (2) a cold-start strategy that departs from SFT-based initialization~\citep{xu2024llava,guo2024mammoth,wei2025advancing,yan2025ad-r1} by directly employing RL, ensuring alignment between reward signals and task objectives from the outset. Training data are organized from easy to hard across multiple stages, with dense and accessible rewards at lower levels supporting effective cold-start training. This staged strategy systematically bridges perception and reasoning within a unified RL framework.

We conduct extensive experiments on \dataset and \datasetplus to evaluate the effectiveness of \method. Results indicate that each component yields measurable gains, with their integration delivering the best overall performance. Moreover, beyond the targeted benchmarks, models trained with \method achieve consistent improvements ($3.47\%$ average) across six benchmarks ~\citep{wu2024vstar,wang2024SpatialEval,li2024seednbench2, wang2025HRBench, masry2022chartqa, chen2024MMStar} covering spatial reasoning, fine-grained visual reasoning, and general tasks, suggesting enhanced general visual perception and reasoning capabilities.

In summary, this work makes the following contributions: \textbf{(1)} We introduce \datasetplus, an extended dataset organized from easy to hard, providing dense supervision for multi-stage RL training; \textbf{(2)} We propose \method, a multi-stage RL framework that integrates cold-start curriculum data (\textit{i.e.}, easy $\rightarrow$ hard) with difficulty-aware detail reward design; \textbf{(3)} Extensive experiments demonstrate that \method not only improves performance on \dataset and \datasetplus but also enhances performance across broader visual benchmarks beyond transit maps. Together, these contributions establish a principled approach to overcoming sparse reward challenges in visual reasoning, advancing the capabilities of MLLMs in structured visual tasks.

\section{Related Work}
\label{sec:related_work}

\textbf{Visual Reasoning in MLLMs.}
The development of MLLMs has rapidly progressed from foundational models that bridge vision and language encoders, such as Flamingo~\citep{alayrac2022flamingo}, to those enhanced by visual instruction tuning like LLaVA~\citep{liu2023llava, liu2024llavanext}. 
To elicit more complex, step-by-step reasoning, researchers have adapted Chain-of-Thought (CoT)~\citep{wei2022chain} prompting from the language domain to the multimodal context (MCoT)~\citep{zhang2023multimodal}.
However, a key limitation of early MCoT is its reliance on purely textual rationales~\citep{wang2025multimodalsurvey}, which can be a bottleneck for expressing fine-grained visual logic~\citep{zhang2025thyme}.
More recent work has thus focused on developing vision-centric reasoning processes~\citep{dong2025insight, man2025argus}, such as generating intermediate visual representations as perception tokens to aid the reasoning chain~\citep{bigverdi2025perceptiontoken}. 
Despite these advances, a significant performance gap persists. 
Benchmarks specifically designed to test abstract, spatial, and logical reasoning, such as VisuLogic~\citep{xu2025visulogic} and \dataset~\citep{feng2025reasonmap}, reveal that even state-of-the-art models struggle with tasks that require high-fidelity visual and topological understanding, underscoring the need for new methods tailored for structured visual domains, such as transit maps.

\begin{figure}[t]
\centering
\includegraphics[width=\linewidth]{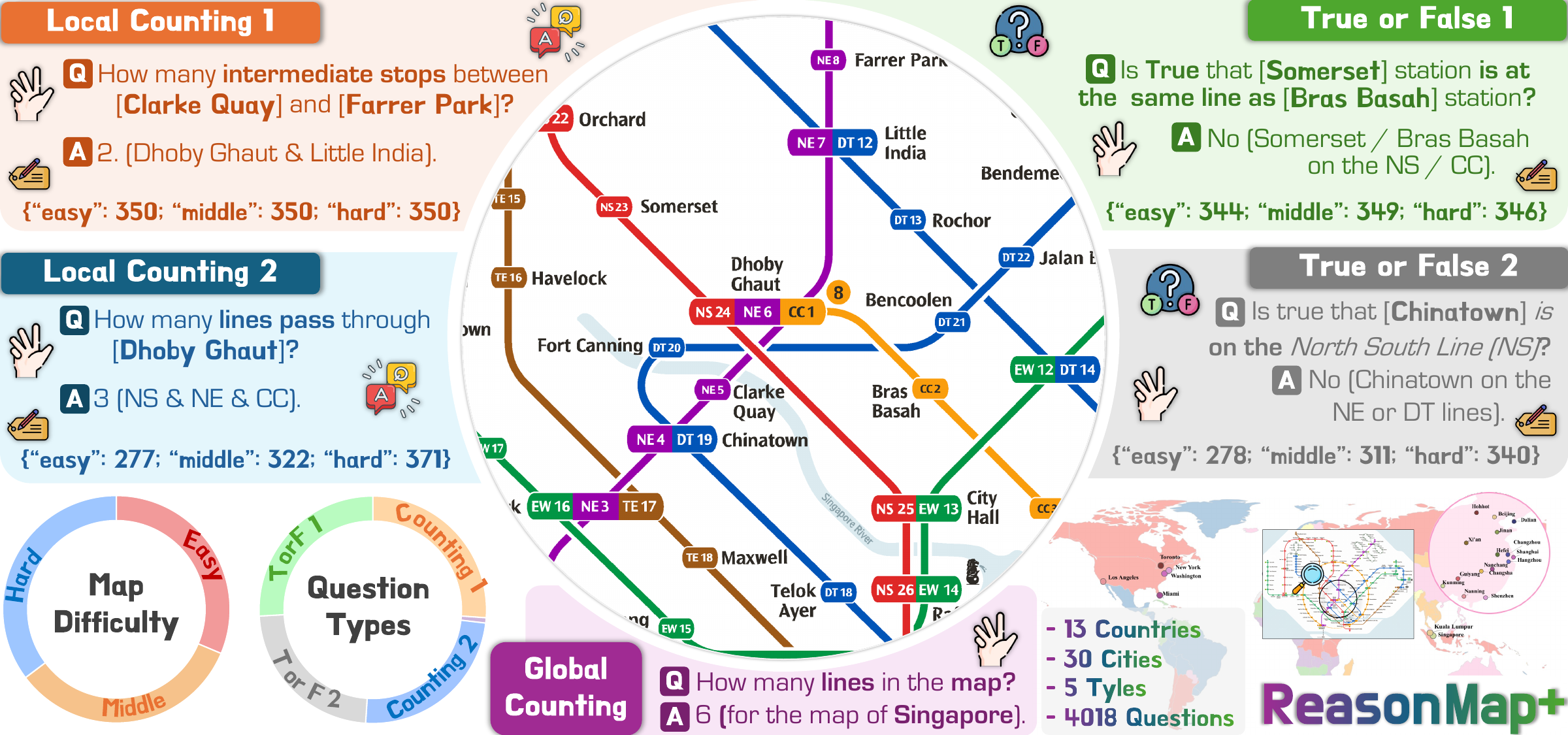} \\
\vspace{-3mm}
\caption{Overview of \datasetplus. \datasetplus comprises $4{,}018$ questions from $5$ extended question types and maps from $30$ cities across $13$ countries.}
\vspace{-1mm}
\label{fig:overview-dataset}
\end{figure}

\textbf{Reinforcement Learning for Reasoning.}
Reinforcement Learning (RL)~\citep{wu2025reinforcement,sarch2025grounded,feng2025efficient} has emerged as a powerful paradigm for improving the reasoning capabilities of multimodal and language-only models beyond the static nature of Supervised Fine-Tuning (SFT)~\citep{liu2023llava,wei2025advancing}. 
The evolution of RL for reasoning spans from traditional RLHF pipelines~\citep{bai2022training} to more stable and direct policy optimization objectives such as Direct Preference Optimization (DPO)~\citep{rafailov2023direct} and Group Relative Policy Optimization (GRPO)~\citep{shao2024deepseekmath}. 
Recent work further demonstrates the effectiveness of structured or curriculum-based RL strategies for enhancing LLM reasoning. 
\revise{Curriculum RL~\citep{parashar2025curriculum} shows that progressing from easy to hard tasks can significantly strengthen reasoning robustness; 
Kimi K1.5~\citep{team2025kimi} adopts multi-stage RL to improve long-chain and tool-integrated reasoning; 
and Logic-RL~\citep{xie2025logic} leverages logical structure to guide policy optimization.} 
While SFT on CoT-rich datasets can activate latent reasoning abilities~\citep{xu2024llava,guo2024mammoth,wei2025advancing}, it often leads to overfitting and cognitive rigidity~\citep{chu2025sft,shen2025vlm,wang2025pixelthink}. 
RL provides a dynamic alternative by allowing models to learn a reasoning policy through exploration and reward~\citep{guo2025deepseekr1,zhang2025surveyrl}. 
However, its direct application to complex visual reasoning remains challenging due to sparse reward signals~\citep{quadros2025llm,cao2024beyond}, where supervision is only provided at the end of long reasoning trajectories~\citep{wang2025harnessing}. 
These limitations motivate the development of an RL framework tailored for structured visual tasks that alleviates sparse rewards from the outset.

\textbf{Spatial Reasoning on Maps.}
Spatial reasoning on transit maps~\citep{wu2020survey} has traditionally been approached with multi-stage computer vision systems. These methods first employ Optical Character Recognition (OCR)~\citep{memon2020handwritten, liu2024llavanext} to extract textual information such as station names, and then use specialized image processing and graph algorithms to identify key topological elements like stations (nodes) and the routes connecting them (edges)~\citep{cherry2006design}. 
Pathfinding is subsequently performed on this extracted graph representation of the transit network using classical search algorithms~\citep{noto2000method, deng2012fuzzy}. 
While logical, these systems are often brittle, as errors in early stages can propagate and lead to incorrect final paths. 
MLLMs offer a promising end-to-end alternative, yet benchmarks consistently show they fail at fundamental spatial tasks like judging relative positions and orientation~\citep{xing2025can}. 
These failures are often attributed to architectural limitations in how vision encoders process positional information~\citep{chen2024solving}. When applied directly to map-based planning, MLLMs struggle to comprehend environmental constraints and execute the multi-hop reasoning required~\citep{feng2025reasonmap}. 
% This has led to hybrid neuro-symbolic systems like LLM-A*, which uses an LLM for high-level planning and a classical A* algorithm for precise, low-level pathfinding. 
Our work addresses above concerns by constructing cold-start data for fine-grained understanding and through multi-stage reinforcement learning training.
\section{\datasetplus Construction}
\label{sec:dataset}

In this section, we first introduce the construction pipeline of \datasetplus as shown in Figure~\ref{fig:overview-dataset}, which builds on \dataset~\citep{feng2025reasonmap}, and comprises three stages: (1) data collection and preprocessing, (2) design-guided construction of question–answer pairs, and (3) quality control. We report a comprehensive statistical overview of \datasetplus in Appendix~\ref{apx:statistical-overview}.

\subsection{Construction Pipeline of \datasetplus}
\label{sec:construction-pipeline}

We follow the construction pipeline of \dataset on stages (1) and (3) for \datasetplus. Specifically, for (1) data collection and preprocessing, we reuse the collected high-resolution transit maps and annotated line-stop information (refer to the Metro Data) in \dataset to drive the following question-answer generation; for (3) quality control, we manually review the automatically generated question-answer pairs to verify correctness and, when necessary, adjust the question distribution to maintain diversity and a balanced difficulty. We then present the details of stage (2).

\textbf{Construction of Question-Answer Pairs in \datasetplus.}
We extend the planning question in \dataset to $5$ related categories covering counting and True or False questions (Figure~\ref{fig:overview-dataset}), preserving consistency between \datasetplus and \dataset. For each category, we construct questions based on predesigned question templates (see Appendix~\ref{apx:question_template}) and automatically derive answers from the Metro Data to form question-answer pairs.

We consider the following question types: (1) \textbf{Global Counting} assesses global fine-grained visual understanding by asking for the number of lines in a map. This type is sparse, and each map yields only one question; (2) \textbf{Local Counting} evaluates local fine-grained visual understanding through two variants: counting the intermediate stops between two specified stops, and counting the number of lines that pass through a specified stop; (3) \textbf{True or False} probes fine-grained visual understanding through two variants: judging the spatial relation between two specified stops, and between one stop and one line. We balance yes and no answers to prevent models from exploiting label frequency.

\textbf{Difficulty Annotation.}
For map difficulty, we follow the manual assignment in \dataset, which uniformly categorizes all maps into three levels (\textit{e.g.}, easy, middle, and hard). For question difficulty, these questions probe basic visual understanding of transit maps rather than the complex visual reasoning required by the planning questions in \dataset. Accordingly, we define question difficulty by the corresponding map difficulty label.

\section{Methodology}
\label{sec:methods}

\begin{figure}[t]
    \centering
    \includegraphics[width=\linewidth]{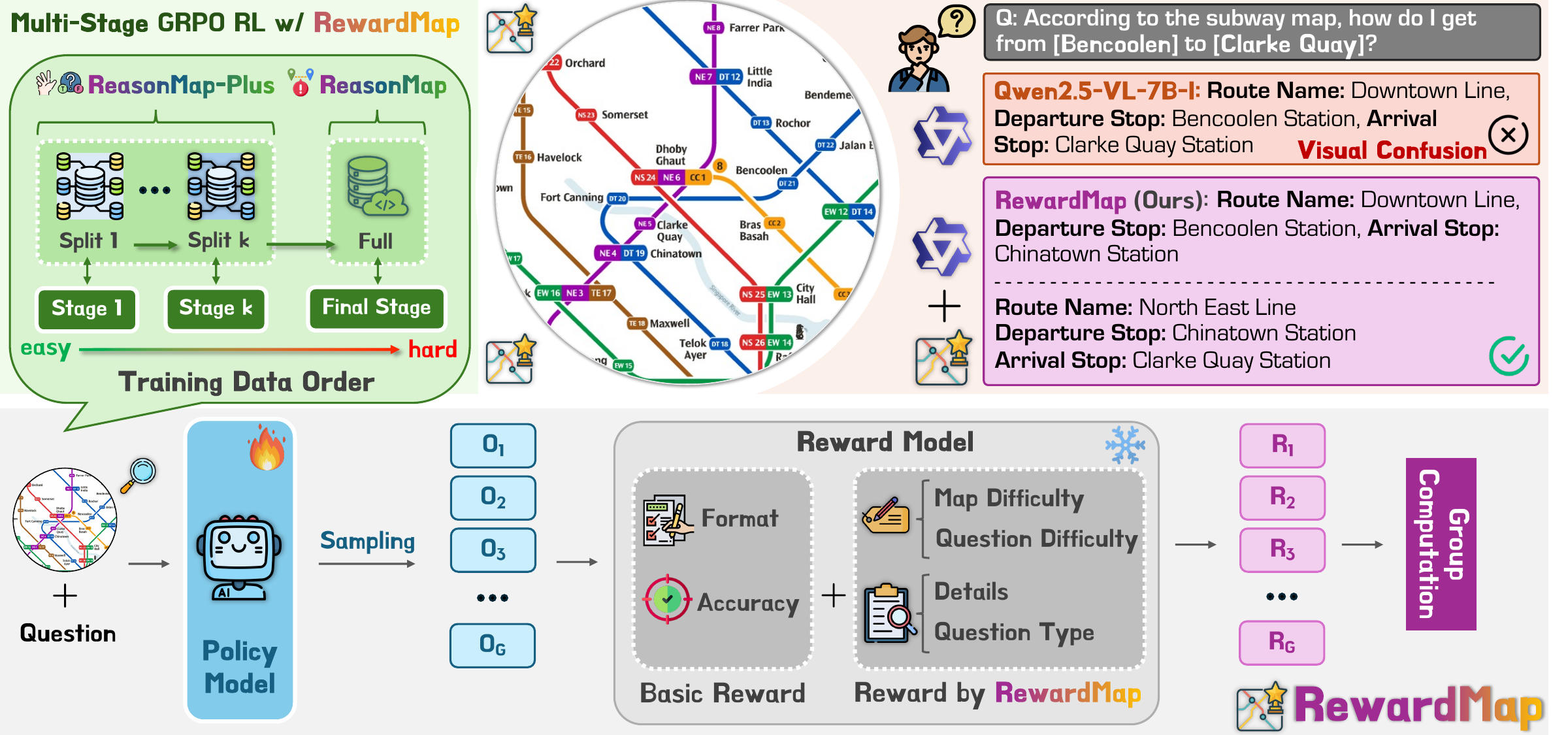} 
    \\
    \vspace{-3mm}
    \caption{Overview of \method. The framework enhances fine-grained visual understanding and reasoning in MLLMs through reinforcement learning with Group Relative Policy Optimization (GRPO). It consists of two key components: (1) a \textbf{difficulty-aware reward design} (Section~\ref{sec:difficulty-aware reward design}), which combines format, correctness, and detail rewards with difficulty-based weighting; and (2) a \textbf{multi-stage RL curriculum} (Section~\ref{sec:multi-stage-rl}), which schedules training data from simple perception tasks to complex reasoning tasks, ensuring effective optimization tackling sparse rewards.}
    \vspace{-1mm}
    \label{fig:overview-methods}
\end{figure}

In this section, we propose \method to enhance fine-grained visual understanding and reasoning in MLLMs. 
We begin by presenting the overview of our target tasks, baseline, and our proposed \method. 
We then introduce the difficulty-aware reward design and illustrate multi-stage reinforcement learning with the Group Relative Policy Optimization (GRPO) process.

\subsection{Overview}
\label{sec:overview}

\textbf{Target Tasks.}
We investigate two target tasks in this paper: (1) fine-grained visual understanding in \datasetplus (Section~\ref{sec:dataset}) and (2) fine-grained visual reasoning in \dataset~\citep{feng2025reasonmap}. Both are cast as Visual Question Answering (VQA), given a high-resolution image $I$ and an instructional question $Q$, the model must produce an answer $A_{\text{formatted}}$ that conforms to the required output format and correctly addresses $Q$. Unlike conventional benchmarks, our tasks foreground fine-grained perception to assess a model’s ability to exploit high-resolution details. Additionally, the route planning task in \dataset further evaluates spatial reasoning.

\textbf{Baseline.}
We conduct baseline experiments under a standard setup with training data from the original \dataset. We train the model with GRPO reinforcement learning with a basic reward function consisting of a format reward and a correctness reward. However, training on this setting exhibits sparse rewards under high task difficulty (see the results of Qwen2.5-VL-3/7B-Instruct in Table~\ref{tab:main}). In GRPO, given an input $x$ and a group $G=\{y_i\}_{i=1}^K$ of sampled outputs with returns $\{r_i\}_{i=1}^K$, the centered group advantage
drives policy updates:
\begin{equation*}
  \hat A_i \;=\; r_i \;-\; \frac{1}{K}\sum_{j=1}^K r_j, 
  \qquad
  \max_{\theta}\;\; \mathcal{L}(\theta) \;=\; 
  \sum_{i=1}^K \hat A_i \,\log \pi_{\theta}(y_i \mid x).
  \label{eq:grpo-adv}
\end{equation*}
With sparse rewards, most $r_i\approx 0$, so $\hat A_i$ either collapses to near zero (all failures) or becomes highly skewed (rare positives), yielding low-signal or high-variance gradients and thus slowing convergence. 

\textbf{\method.}
To mitigate the sparse reward issue observed in the  baseline, we propose \method, which consists of two components: (1) a detail-oriented reward with difficulty-aware weighting, and (2) a multi-stage RL regimen that exploits dense-reward questions from \datasetplus for effective cold start. Implementation details are provided below.

\subsection{Difficulty-aware Reward Design}
\label{sec:difficulty-aware reward design}

Our reward function comprises three terms: format reward, correctness reward, and detail reward, scaled by a difficulty factor and a weighting coefficient for the detail reward:
\begin{equation*}
  R \;=\; W_{\text{difficulty}}\!\left( R_{\text{format}} \;+\; R_{\text{correctness}} \;+\; \alpha \times R_{\text{detail}} \right),
  \label{eq:whole-reward}
\end{equation*}
where \(R_{\text{format}}\), \(R_{\text{correctness}}\), and \(R_{\text{detail}}\) denote the format, correctness, and detail rewards, respectively; \(\alpha>0\) controls the relative strength of the detail term (\(\alpha>0\) is set to $0.5$ for subsequent training); and \(W_{\text{difficulty}}>0\) scales the overall reward according to difficulty.

\textbf{Format Reward.} 
The format reward enforces compliance with task-specific output conventions: for \datasetplus, answers are elicited within a ``\verb|\boxed{}|'' to localize the output; for \dataset, the reward is computed using the benchmark’s original formatting specification\footnote{Please refer to Appendix~A.1 in \dataset paper for more details.}.

% % \vspace{0.5mm}
% \begin{tcolorbox}[title=Planning Question Format in \dataset,colback=blue!5!white,colframe=blue!60!black,fonttitle=\bfseries]
% % According to the subway map, how do I get from xx to xx? The format should be strictly followed:
% \begin{verbatim}
% Route Name: Line x
% Departure Stop: xx Station
% Arrival Stop: xx Station
% --
% ...
% \end{verbatim}
% \end{tcolorbox}

\textbf{Correctness Reward.}
For training data in \datasetplus, we use exact-match scoring to compute rewards, as these questions have a single deterministic ground truth (\textit{e.g.}, numerals or yes/no). For training data in our proposed \dataset, the correctness reward is computed using the benchmark’s official evaluation algorithm\footnote{See Appendix~B.1 of the \dataset paper~\citep{feng2025reasonmap}.}.

\textbf{Detail Reward.}
To alleviate sparse rewards caused by the difficulty of planning tasks, we add a detail reward that grants partial credit for correct items of the answer. Specifically, we reward/penalize correctness of the origin and destination stops, route names, transfer stations, and the number of route segments (see the computation pipeline in Algorithm~\ref{alg:detail-reward}). Additionally, we modulate this influence of detail reward on training with a weighting coefficient $\alpha$.

\textbf{Difficulty-Aware Weighting.}
To incorporate problem difficulty, we scale the sum of the three rewards by a difficulty-aware weight. We consider two fine-grained factors: (1) Map difficulty (for all training data from \dataset and \datasetplus), with weights assigned by the three levels (\textit{e.g.}, easy, medium, hard, see Appendix~\ref{apx:statistical-overview}); and (2) Question difficulty (for the training data in \dataset), with weights determined by the transfer count of the required route. 
The weighting scheme is defined as follows:
\begin{equation*}
  W_{\text{difficulty}}
  \;=\;
  W_{\text{map}} \;+\; W_{\text{question}},
\end{equation*}
\begin{equation*}
  W_{\text{map}} \;=\;
  \begin{cases}
    \gamma_e, & \text{map difficulty} = \text{easy}\\
    \gamma_m, & \text{map difficulty} = \text{medium}\quad,\\
    \gamma_h, & \text{map difficulty} = \text{hard}
  \end{cases}
  \qquad
  W_{\text{question}} \;=\;
  \begin{cases}
    \beta_0, & \text{transfer count} = \text{0}\\
    \beta_1, & \text{transfer count} \geq \text{1}
  \end{cases}
    \quad .
\end{equation*}

\subsection{Multi-Stage GRPO-based Reinforcement Learning}
\label{sec:multi-stage-rl}

% To fully exploit \datasetplus, we introduce a multi-stage GRPO-based RL curriculum that leverages dense reward questions. 
To effectively exploit \datasetplus and alleviate the sparse-reward issue in complex tasks such as route planning in \dataset, 
we design a \textbf{multi-stage curriculum} built upon GRPO-based reinforcement learning. 
% The core idea is to control the ordering of training data throughout RL to mitigate sparse reward effects from difficult tasks like route planning in \dataset. 
The core idea is to progressively schedule training data in a principled manner, ensuring a smoother optimization process and more stable reward propagation. 
We adhere to two complementary principles:
% (1) \textbf{Global principle} impose an easy to hard curriculum by partitioning training data from \datasetplus and \dataset into stages by question type (\textit{i.e.}, True or False $\rightarrow$ counting $\rightarrow$ planning) and target task (\textit{i.e.}, fine-grained visual understanding $\rightarrow$ fine-grained visual reasoning), as shown in Figure~\ref{fig:overview-methods}; (
% 2) \textbf{Local principle} inject randomness within each stage by shuffling samples rather than strictly sorting by difficulty metrics (\textit{e.g.}, map difficulty or question difficulty).

\noindent\textbf{(1) Global curriculum principle.} 
We impose a coarse-to-fine learning schedule by partitioning tasks into distinct stages according to both \emph{question type} (from binary judgment $\rightarrow$ counting $\rightarrow$ planning) and \emph{target task} (from fine-grained visual understanding $\rightarrow$ fine-grained visual reasoning). 
This global ordering ensures that the agent acquires fundamental perceptual skills before engaging in more abstract reasoning, as illustrated in Figure~\ref{fig:overview-methods}.

\noindent\textbf{(2) Local stochasticity principle.} 
Within each stage, we avoid strict deterministic ordering by introducing randomness, \textit{i.e.}, shuffling training samples instead of ranking them solely by heuristic difficulty metrics (\textit{e.g.}, map or question complexity). 
This stochasticity prevents overfitting to a fixed curriculum trajectory and enhances robustness.
% while still preserving the overall easy-to-hard progression.

By jointly applying these principles, the proposed multi-stage RL scheme transforms curriculum training into a structured combination of \emph{reward shaping} and \emph{task scheduling}, thereby enabling effective reinforcement learning on inherently sparse-reward visual reasoning problems.

\section{Experiments}
\label{sec:exps}

\subsection{Experimental Setups}
\label{sec:experimental-setup}

\begin{table*}[t]
\centering
\vspace{-2mm}
\caption{Evaluations of reference models and fine-tuned models on \dataset and \datasetplus. ``$S.$'' represents results for short questions, while ``$L.$'' denotes results for long questions. \textbf{Bold} indicates the best results among fine-tuned models, while \underline{underline} represents the second best.}
\label{tab:main}
\resizebox{\linewidth}{!}{
\setlength{\tabcolsep}{0.5mm}
\begin{tabular}{llcccc}
\hline
\multirow{2}{*}{\textbf{Model}} 
& \multirow{2}{*}{\textbf{Training Data}} 
& \multicolumn{2}{c}{\textbf{\dataset ($S. / L.$)}} 
& \multicolumn{2}{c}{\textbf{\datasetplus}}\\ 
& &  Weighted Acc. & Weighted Map Score 
& Weighted Acc. & Weighted Acc. (Count / TorF)  \\ 
\midrule\midrule
\addlinespace
\multicolumn{6}{c}{\textit{Reference Models}} \\
\midrule
Kimi-VL-A3B-Thinking & - & 
$5.47\%$ / $5.47\%$  & $2.44$ / $3.17$ & 
$33.95\%$ & $18.17\%$ / $50.39\%$ \\
Kimi-VL-A3B-Instruct & - & 
$12.76\%$ / $12.33\%$  & $3.30$ / $5.37$ & 
$32.55\%$ & $14.75\%$ / $51.08\%$ \\
Qwen2.5-VL-3B-Instruct & - & 
$8.68\%$ / $7.99\%$  & $2.75$ / $3.70$ & 
$37.61\%$ & $22.68\%$ / $53.16\%$  \\
Qwen2.5-VL-32B-Instruct & - & 
$16.49\%$ / $15.71\%$  & $3.88$ / $6.84$ & 
$58.32\%$ & $46.96\%$ / $70.14\%$ \\
Qwen2.5-VL-72B-Instruct & - & 
$26.65\%$ / $24.22\%$  & $5.09$ / $8.80$ & 
$53.21\%$ & $43.46\%$ / $63.36\%$ \\ % TODO
\hdashline
Seed1.5-VL & - & %  (id: `doubao-115')
$34.20\%$ / $38.02\%$  & $5.25$ / $11.96$ & 
$73.58\%$ & $65.26\%$ / $82.23\%$ \\
\revise{GPT-4o} & \revise{-} &
\revise{$41.15\%$ / $42.80\%$}  & \revise{$6.84$ / $13.57$} & 
\revise{$64.42\%$} & \revise{$59.28\%$ / $69.77\%$} \\
\revise{GPT-5} & \revise{-} &
\revise{$59.98\%$ / $62.50\%$}  & \revise{$9.48$ / $19.75$} & 
\revise{$88.95\%$} & \revise{$86.40\%$ / $91.60\%$} \\
\midrule
\addlinespace
\multicolumn{6}{c}{\textit{Baseline \& \method}} \\
\midrule
Qwen2.5-VL-7B-Instruct  & - & 
$13.28\%$ / $7.12\%$ & $4.01$ / $5.74$ & 
$44.21\%$ & $37.39\%$ / $51.32\%$ \\ 
\hdashline
+ RL (GRPO) & $R_{train}$ & 
$26.22\%$ / $26.04\%$ & $5.52$ / $9.52$ & 
$44.64\%$ & $37.57\%$ / $52.01\%$ \\ % low sampling baseline
\revise{+ RL (REINFORCE++)} & \revise{$R_{train}$} & 
\revise{$27.17\%$ / $27.60\%$} & \revise{$5.68$ / $10.12$} & 
\revise{$44.64\%$} & \revise{$36.82\%$ / $52.79\%$} \\
\revise{+ RL (ReMax)} & \revise{$R_{train}$} & 
\revise{$26.22\%$ / $27.26\%$} & \revise{$5.57$ / $9.99$} & 
\revise{$45.39\%$} & \revise{$38.37\%$ / $52.70\%$} \\
% + RL + Reward Design & $R_{train}$ & 
% 29.08\% / 29.95\% & $5.88$ / $10.53$ & 
% $45.16\%$ & $37.79\%$ / $52.84\%$ \\ % difficulty aware (1024w) 用于消融(这里不提)
+ SFT    & $RPlus_{train}$ & 
$13.63\%$ / $9.11\%$ & $4.09$ / $6.25$ & 
$57.93\%$ & $50.73\%$ / $65.44\%$ \\ 
+ SFT $\rightarrow$ RL     & $RPlus_{train}$ + $R_{train}$ & 
 $28.82\%$ / $\underline{30.38\%}$ & $5.88$ / $\underline{10.62}$ & 
 $60.53\%$ & $55.38\%$ / $65.90\%$ \\ % low sampling baseline
+ RL (baseline) & $RPlus_{train}$ + $R_{train}$ &
$\underline{29.51\%}$ / $29.51\%$ & $\underline{6.00}$ / $10.41$ & 
$\underline{67.61\%}$ & $\underline{68.37\%}$ / $\underline{66.82\%}$ \\ % baseline RL
% + RL + Multi-Stage Design & $RPlus_{train}$ + $R_{train}$ &
% $30.64\%$ / $31.51\%$ & $6.08$ / $10.88$ & 
% $73.12\%$ & $69.52\%$ / $76.88\%$ \\ % 都用于消融(这里不提) difficultly-aware \& curriculum (planning shuffle)
% + RL + Reward Design & $RPlus_{train}$ + $R_{train}$ &
% $30.56\%$ / $30.38\%$ & $6.12$ / $10.62$ & 
% $71.07\%$ & $67.61\%$ / $74.67\%$ \\ % 都用于消融(这里不提)
+ \method & $RPlus_{train}$ + $R_{train}$ &
$\mathbf{31.51\%}$ / $\mathbf{31.77\%}$ & $\mathbf{6.21}$ / $\mathbf{11.22}$ & 
$\mathbf{74.25\%}$ & $\mathbf{72.18\%}$ / $\mathbf{76.42\%}$ \\
\hline
\end{tabular}}
\vspace{-4mm}
\end{table*}

\textbf{Training Details.}
We conduct training experiments under various settings with the training data from \dataset ($696$ samples, $R_{train}$)~\citep{feng2025reasonmap} and our proposed \datasetplus ($2,570$ samples, $RPlus_{train}$) on Qwen2.5-VL models~\citep{bai2025qwen25} with $8$ NVIDIA H800 GPUs. For GRPO~\citep{shao2024deepseekmath} RL training, we use AdamW with an initial learning rate of $1.0 \times 10^{-6}$ and a KL divergence coefficient of $1.0 \times 10^{-3}$. The global batch size is $16$, and we sample $8$ responses per query. Besides RL training, we conduct baseline experiments using SFT with training data from \datasetplus. We apply LoRA~\citep{hu2022lora} to the language blocks and the text-vision projector with an initial learning rate of $1.0\times10^{-4}$. For implementation, we adopt LLaMA-Factory~\citep{zheng2024llamafactory} and VeRL~\citep{sheng2024hybridflow}. \revise{Additionally, we include two RL training baselines with REINFORCE++~\citep{hu2025reinforce++} and ReMax~\citep{li2023remax}.} 
% All experiments were conducted on $8$ NVIDIA H800 GPUs. 

\textbf{Inference Details.}
All evaluations use greedy decoding (temperature=$0$). For open-source models, we cap the maximum output length at $2,048$ tokens and retain all other settings from the official HuggingFace configurations; deployments use PyTorch with Transformers\footnote{\url{https://github.com/huggingface/transformers}} on $8$ NVIDIA H800 GPUs. For closed-source models, we evaluate through official APIs using default settings. The evaluated models include: Kimi-VL-A3B-Insturct/Thinking~\citep{team2025kimi}, Qwen2.5-VL-3/7/32/72B-Instruct~\citep{bai2025qwen25}, Seed1.5-VL~\citep{guo2025seed1}, \revise{GPT-5/4o~\citep{gpt4o}}.

\textbf{Evaluation Datasets.}
We first evaluate on the test sets of \dataset and \datasetplus, and we report the metrics adjusted by the difficulty-aware weighting scheme (see details in Appendix~\ref{apx:weighting-details}). To further assess the capability gains brought by \method, we next employ six widely-used benchmarks spanning three dimensions (\textit{e.g.}, spatial reasoning, fine-grained visual reasoning, and general tasks): SEED-Bench-2-Plus~\citep{li2024seednbench2}, SpatialEval~\citep{wang2024SpatialEval}, V$^{*}$Bench~\cite{wu2024vstar}, HRBench~\citep{wang2025HRBench}, ChartQA~\citep{masry2022chartqa}, and MMStar~\citep{chen2024MMStar} (see Appendix~\ref{apx:evaluation-datasets-details} for details). All evaluations are conducted with VLMEvalKit\footnote{\url{https://github.com/open-compass/VLMEvalKit}}.

\subsection{Main Results}

\textbf{Results on \dataset and \datasetplus.}
We evaluate the proposed \method on \dataset and \datasetplus, both of which provide fine-grained difficulty annotations to assess visual understanding, visual reasoning, and spatial reasoning. In addition to the baselines introduced in Section~\ref{sec:overview}, we compare against two widely-used settings: (1) \textbf{SFT $\rightarrow$ RL baseline}, which applies SFT on \datasetplus followed by RL on \dataset, and (2) \textbf{RL baseline}, which performs RL on the combined training data from \dataset and \datasetplus. For both baselines, the reward design includes only format and correctness terms.

As shown in Table~\ref{tab:main}, \method consistently outperforms all baselines across different question templates in \dataset and question types of \datasetplus. On \dataset, it substantially surpasses the best open-source result (Qwen2.5-VL-72B-Instruct) and approaches the performance of the closed-source model (Seed1.5-VL). On \datasetplus, \method not only exceeds all open-source models but also outperforms Seed1.5-VL.

\begin{table*}[t]
\centering
\vspace{-2mm}
\caption{Evaluation of reference models and fine-tuned models on various benchmarks. \textbf{Bold} indicates the best results among fine-tuned models, while \underline{underline} represents the second best. 
$\dagger, \ddagger, \$, *, \S$ denote the results from the technical report or the official HuggingFace repository (see result sources in Appendix~\ref{apx:result-source}), while all other results are obtained from our own experiments.}
\label{tab:main-multi-benchmark}
\resizebox{\linewidth}{!}{
\setlength{\tabcolsep}{0.5mm}
\begin{tabular}{lccccccc}
\toprule
\multirow{2}{*}{\textbf{Model}} &
\multicolumn{2}{c}{\textbf{Spatial Reasoning}} &
\multicolumn{2}{c}{\textbf{Fine-grained Visual Reasoning}} &
\multicolumn{2}{c}{\textbf{General Task}} &
\multirow{2}{*}{\textbf{Avg.}}\\
 & \textbf{SEED-Bench-2-Plus} & \textbf{SpatialEval} & $\mathbf{V^{*}}$ & \textbf{HRBench} & \textbf{ChartQA} & \textbf{MMStar} \\
\midrule
\midrule
\addlinespace
\multicolumn{8}{c}{\textit{Reference Models}} \\
\midrule
% Kimi-VL-A3B-Thinking 
%  & & &  & & & $70.40\%^*$ \\
Kimi-VL-A3B-Instruct 
 & $58.49\%$ & $52.64\%$ & $59.16\%$ & $55.38\%$ & $87.08\%$ & $61.70\%^*$ & $62.41\%$\\
Qwen2.5-VL-3B-Instruct 
 & $58.86\%$ & $55.04\%$ & $58.12\%$ & $66.25\%$ & $84.00\%^\ddagger$ & $55.90\%^\ddagger$ & $63.03\%$ \\
Qwen2.5-VL-32B-Instruct 
 & $72.10\%^\$$ & $57.99\%$ & $82.72\%$ & $63.50\%$ & $64.90\%^\$$ & $69.50\%^\S$ & $68.45\%$ \\
Qwen2.5-VL-72B-Instruct 
& $73.00\%^\$$ & - & $86.40\%^\dagger$ & - & $89.50\%^\ddagger$ & $70.80\%^\ddagger$ & - \\
\hdashline 
Seed1.5-VL 
& - & - & $89.00\%^\dagger$ & - & $87.40\%^\dagger$ & $76.20\%^\dagger$ & - \\
\midrule
\addlinespace
\multicolumn{8}{c}{\textit{Baseline \& \method}} \\
\midrule
Qwen2.5-VL-7B-Instruct & $60.97\%$ &  $57.30\%$ & $\underline{78.01\%}$ & $68.75\%$ & $86.12\%$ & $61.67\%$ & $68.80\%$\\
+ SFT $\rightarrow$ RL & \resup{$\underline{61.59\%}$}{$0.62\%$} & \resup{$\underline{69.06\%}$}{$11.76\%$} & \resup{$\underline{78.01\%}$}{$0.00\%$} & \resup{$\underline{71.00\%}$}{$2.25\%$} & \resup{$\underline{86.92\%}$}{$0.80\%$} & \resup{$\mathbf{62.27\%}$}{$0.60\%$} & \resup{$\underline{71.48\%}$}{$2.68\%$} \\
+ \method & \resup{$\mathbf{61.96\%}$}{$0.99\%$} & \resup{$\mathbf{70.81\%}$}{$13.51\%$} & \resup{$\mathbf{80.10\%}$}{$2.09\%$} & \resup{$\mathbf{71.25\%}$}{$2.50\%$} & \resup{$\mathbf{87.24\%}$}{$1.12\%$} & \resup{$\mathbf{62.27\%}$}{$0.60\%$} & \resup{$\mathbf{72.27\%}$}{$3.47\%$} \\
\bottomrule
\end{tabular}}
\vspace{-4mm}
\end{table*}

\textbf{Results on Other Benchmarks.}
We further evaluate the generalization ability of our method on six benchmarks spanning three task categories introduced in Section~\ref{sec:experimental-setup}. Table~\ref{tab:main-multi-benchmark} reports the results of reference models, the SFT $\rightarrow$ RL baseline, and our proposed \method. Across all six benchmarks, \method achieves consistent improvements, with the most substantial gain of $13.51\%$ observed on SpatialEval. While the SFT $\rightarrow$ RL baseline also yields stable improvements, its performance remains inferior to \method. These results highlight the significant contributions of both \method and the datasets (both \dataset and \datasetplus) to enhancing model general capability. \revise{Beyond the aggregate results, we further report task-wise improvements. On SEED-Bench-2-Plus, \method improves the map split. For SpatialEval, accuracy rises substantially on \textit{mazenav} ($19.60\%\rightarrow57.20\%$) and moderately on \textit{spatialreal} ($70.37\%\rightarrow72.59\%$). On HRBench, performance increases for both single-image reasoning ($85.25\%\rightarrow88.00\%$) and cross-image alignment ($52.25\%\rightarrow54.00\%$). For MMStar, \method achieves steady gains across fine-grained perception ($59.60\%\rightarrow60.80\%$), instance reasoning ($68.00\%\rightarrow71.20\%$), and mathematical reasoning ($59.60\%\rightarrow63.20\%$). These results show that \method consistently strengthens visual grounding and robustness in diverse spatial and reasoning tasks.}

\subsection{Qualitative Results}
\label{sec:qualitative-results}

\begin{figure}[t]
    \centering
    \includegraphics[width=1.0\linewidth]{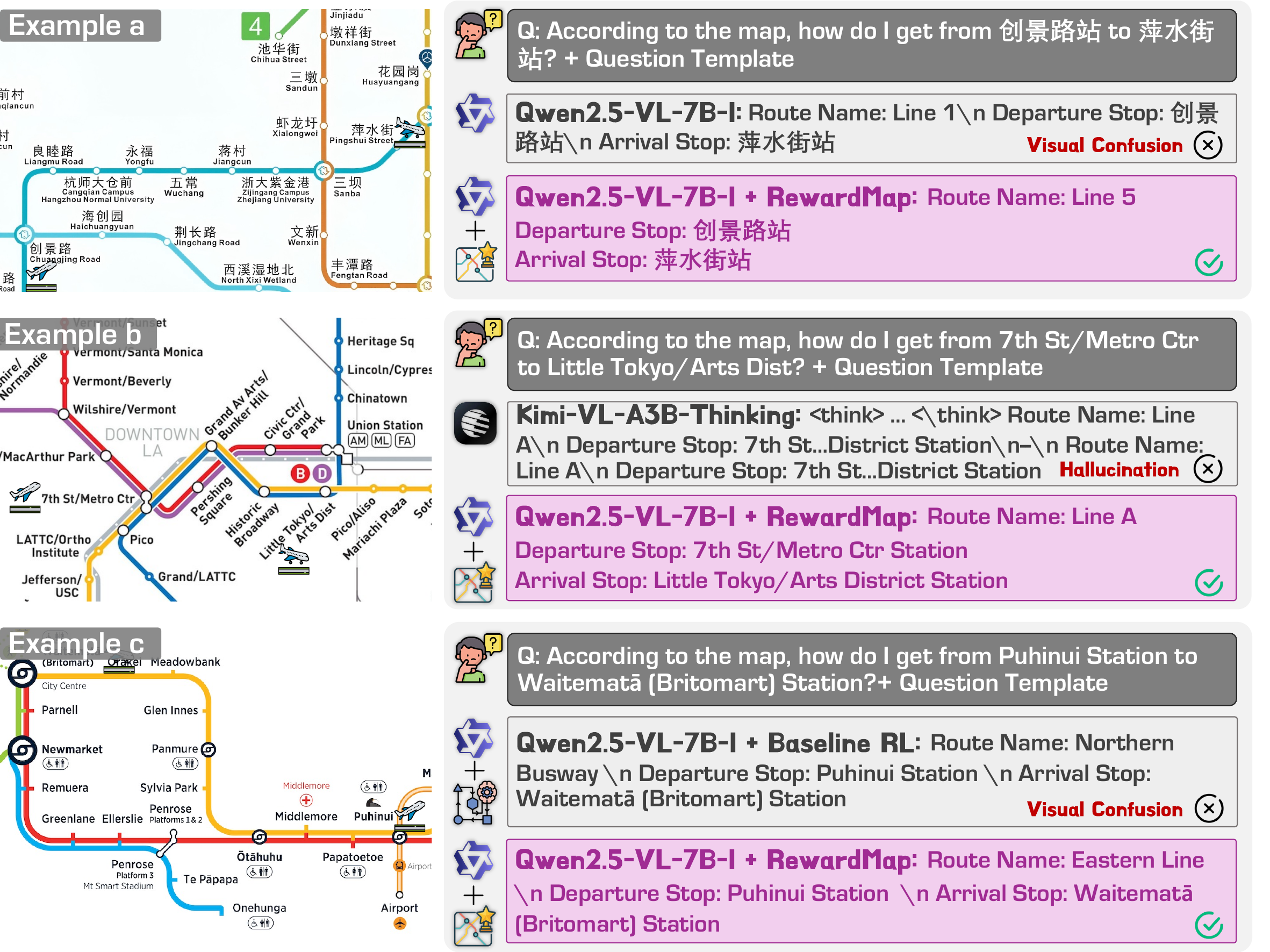} 
    \\
    \vspace{-4.5mm}
    \caption{Qualitative comparisons among reference models, baseline, and our proposed \method. We crop and zoom in on the transit map for clearer presentation.}
    %  Baseline RL denotes the result of the fine-tuned model with training data only from \dataset.
    \vspace{-5mm}
    \label{fig:qualitative-results}
\end{figure}

Figure~\ref{fig:qualitative-results} presents a qualitative comparison between reference models, the baseline RL model, and our proposed \method. Across diverse maps, we observe that reference models and the baseline RL model often suffer from visual confusion (\textit{e.g.}, mistaking the route or stop as shown by Example a \& c in Figure~\ref{fig:qualitative-results}) or even hallucination (\textit{e.g.}, repeating the same route many times in Figure~\ref{fig:qualitative-results} (b)). In contrast, \method consistently produces the correct target routes. These examples highlight the effectiveness of \method in handling visually complex maps (\textit{e.g.}, Example a \& b in Figure~\ref{fig:qualitative-results}) and reducing both visual confusions and hallucinations. \revise{Additionally, we present further comparison cases that provide additional evidence that \method improves visual grounding and reduces the likelihood of visual confusion or hallucination. These examples are included in Appendix~\ref{apx:comparison-cases}.}

\begin{table*}[t]
\centering
\vspace{-2mm}
\caption{Ablation on reward design and multi-stage design of \method. ``$S.$'' represents results for short questions, while ``$L.$'' denotes results for long questions.}
\label{tab:ablation-rewardmap}
\resizebox{\linewidth}{!}{
\setlength{\tabcolsep}{0.5mm}
\begin{tabular}{llcccc}
\hline
\multirow{2}{*}{\textbf{Model}} 
& \multirow{2}{*}{\textbf{Training Data}} 
& \multicolumn{2}{c}{\textbf{\dataset ($S. / L.$)}} 
& \multicolumn{2}{c}{\textbf{\datasetplus}}\\ 
& &  Weighted Acc. & Weighted Map Score 
& Weighted Acc. & Weighted Acc. (Count / TorF)  \\ 
\midrule\midrule
RL (baseline) & $R_{train}$ & 
$26.22\%$ / $26.04\%$ & $5.52$ / $9.52$ & 
$44.64\%$ & $37.57\%$ / $52.01\%$ \\ % low sampling baseline
RL + Reward Design & $R_{train}$ & 
29.08\% / 29.95\% & $5.88$ / $10.53$ & 
$45.16\%$ & $37.79\%$ / $52.84\%$ \\ % difficulty aware (1024w) 用于消融
\hdashline
RL (baseline) & $RPlus_{train}$ + $R_{train}$ &
$29.51\%$ / $29.51\%$ & $6.00$ / $10.41$ & 
$67.61\%$ & $68.37\%$ / $66.82\%$ \\ % baseline RL
RL + Reward Design & $RPlus_{train}$ + $R_{train}$ &
$30.56\%$ / $30.38\%$ & $6.12$ / $10.62$ & 
$71.07\%$ & $67.61\%$ / $74.67\%$ \\ 
RL + Multi-Stage Design & $RPlus_{train}$ + $R_{train}$ &
$30.64\%$ / $31.51\%$ & $6.08$ / $10.88$ & 
$73.12\%$ & $69.52\%$ / $76.88\%$ \\ % 用于消融 difficultly-aware \& curriculum (planning shuffle)
\method & $RPlus_{train}$ + $R_{train}$ &
$\mathbf{31.51\%}$ / $\mathbf{31.77\%}$ & $\mathbf{6.21}$ / $\mathbf{11.22}$ & 
$\mathbf{74.25\%}$ & $\mathbf{72.18\%}$ / $\mathbf{76.42\%}$ \\
\hline
\end{tabular}}
\vspace{-4mm}
\end{table*}

\begin{table*}[t]
\centering
\vspace{-2mm}
\caption{Ablation on \datasetplus. ``$S.$'' represents results for short questions, while ``$L.$'' denotes results for long questions.}
\label{tab:ablation-datasetplus}
\resizebox{\linewidth}{!}{
\setlength{\tabcolsep}{0.5mm}
\begin{tabular}{llcccc}
\hline
\multirow{2}{*}{\textbf{Model}} 
& \multirow{2}{*}{\textbf{Training Data}} 
& \multicolumn{2}{c}{\textbf{\dataset ($S. / L.$)}} 
& \multicolumn{2}{c}{\textbf{\datasetplus}}\\ 
& &  Weighted Acc. & Weighted Map Score 
& Weighted Acc. & Weighted Acc. (Count / TorF)  \\ 
\midrule\midrule
RL (baseline) & $R_{train}$ & 
$26.22\%$ / $26.04\%$ & $5.52$ / $9.52$ & 
$44.64\%$ & $37.57\%$ / $52.01\%$ \\ % low sampling baseline
RL (baseline) & $RPlus_{train}$ + $R_{train}$ &
$29.51\%$ / $29.51\%$ & $6.00$ / $10.41$ & 
$67.61\%$ & $68.37\%$ / $66.82\%$ \\ % baseline RL
SFT $\rightarrow$ RL     & $RPlus_{train}$ + $R_{train}$ & 
 $28.82\%$ / $30.38\%$ & $5.88$ / $10.62$ & 
 $60.53\%$ & $55.38\%$ / $65.90\%$ \\ % low sampling baseline
\hline
\end{tabular}}
\vspace{-4mm}
\end{table*}

\begin{table*}[t]
\centering
\vspace{-2mm}
\caption{Ablation on granularity of multi-stage design. ``$S.$'' represents results for short questions, while ``$L.$'' denotes results for long questions.}
\label{tab:ablation-granularity}
\resizebox{\linewidth}{!}{
\setlength{\tabcolsep}{0.5mm}
\begin{tabular}{llcccc}
\hline
\multirow{2}{*}{\textbf{Model}} 
& \multirow{2}{*}{\textbf{Training Data}} 
& \multicolumn{2}{c}{\textbf{\dataset ($S. / L.$)}} 
& \multicolumn{2}{c}{\textbf{\datasetplus}}\\ 
& &  Weighted Acc. & Weighted Map Score 
& Weighted Acc. & Weighted Acc. (Count / TorF)  \\ 
\midrule\midrule
RL (baseline) & $RPlus_{train}$ + $R_{train}$ &
$29.51\%$ / $29.51\%$ & $6.00$ / $10.41$ & 
$67.61\%$ & $68.37\%$ / $66.82\%$ \\ % baseline RL
RL (coarse-grained Multi-Stage) & $RPlus_{train}$ + $R_{train}$ &
$29.60\%$ / $30.21\%$ & $5.98$ / $10.83$ &
$70.30\%$ & $66.02\%$ / $74.76\%$ \\
RL (Multi-Stage) = \method & $RPlus_{train}$ + $R_{train}$ &
$31.51\%$ / $31.77\%$ & $6.21$ / $11.22$ & 
$74.25\%$ & $72.18\%$ / $76.42\%$ \\
\hline
\end{tabular}}
\vspace{-4mm}
\end{table*}

\begin{table*}[t]
\centering
\caption{\revise{Ablation on the hyperparameter $\alpha$. ``$S.$'' represents results for short questions, while ``$L.$'' denotes results for long questions.}}
\label{tab:ablation-alpha}
\resizebox{\linewidth}{!}{
\setlength{\tabcolsep}{0.5mm}
\begin{tabular}{llcccc}
\hline
\multirow{2}{*}{\textbf{Model}} 
& \multirow{2}{*}{\textbf{Training Data}} 
& \multicolumn{2}{c}{\textbf{\dataset ($S. / L.$)}} 
& \multicolumn{2}{c}{\textbf{\datasetplus}}\\ 
& &  Weighted Acc. & Weighted Map Score 
& Weighted Acc. & Weighted Acc. (Count / TorF)  \\ 
\midrule\midrule
\revise{+ \method [$\alpha$=0.3]} & \revise{$RPlus_{train}$ + $R_{train}$} &
\revise{$30.73\%$ / $31.16\%$} & \revise{$6.01$ / $10.81$} & 
\revise{$72.08\%$} & \revise{$68.41\%$ / $75.91\%$} \\
\revise{+ \method [$\alpha$=0.5]} & \revise{$RPlus_{train}$ + $R_{train}$} &
\revise{$31.51\%$ / $31.77\%$} & \revise{$6.21$ / $11.22$} & 
\revise{$74.25\%$} & \revise{$72.18\%$ / $76.42\%$} \\
\revise{+ \method [$\alpha$=0.7]} & \revise{$RPlus_{train}$ + $R_{train}$} &
\revise{$32.03\%$ / $32.20\%$} & \revise{$6.23$ / $11.20$} &
\revise{$72.81\%$} & \revise{$70.14\%$ / $75.59\%$} \\
\hline
\end{tabular}}
\vspace{-4mm}
\end{table*}

\begin{table*}[!h]
\centering
\vspace{-2mm}
\caption{\revise{Ablation on the difficulty-aware weights. ``$S.$'' represents results for short questions, while ``$L.$'' denotes results for long questions.}}
\label{tab:ablation-diff-weights}
\resizebox{\linewidth}{!}{
\setlength{\tabcolsep}{0.1mm}
\begin{tabular}{llcccc}
\hline
\multirow{2}{*}{\textbf{Model}}
& \multirow{2}{*}{\textbf{Training Data}}
& \multicolumn{2}{c}{\textbf{\dataset ($S. / L.$)}}
& \multicolumn{2}{c}{\textbf{\datasetplus}} \\
& & Weighted Acc. & Weighted Map Score 
& Weighted Acc. & Weighted Acc. (Count / TorF) \\
\midrule\midrule
Qwen2.5-VL-7B-Instruct & - &
$13.28\%$ / $7.12\%$ & $4.01$ / $5.74$ &
$44.21\%$ & $37.39\%$ / $51.32\%$ \\
\revise{+ Reward Design [$\gamma_{e/m/h}=(1.0,1.1,1.2)$]} & \revise{$R_{train}$} &
\revise{$26.91\%$ / $28.47\%$} & \revise{$5.64$ / $10.24$} &
\revise{$45.34\%$} & \revise{$37.97\%$ / $53.02\%$} \\
\revise{+ Reward Design [$\gamma_{e/m/h}=(1.0,1.2,1.5)$]} & \revise{$R_{train}$} &
\revise{$29.08\%$ / $29.95\%$} & \revise{$5.88$ / $10.53$} &
\revise{$45.16\%$} & \revise{$37.79\%$ / $52.84\%$} \\
\revise{+ Reward Design [$\gamma_{e/m/h}=(1.0,1.5,2.0)$]} & \revise{$R_{train}$} &
\revise{$29.86\%$ / $28.99\%$} & \revise{$5.93$ / $10.31$} &
\revise{$45.16\%$} & \revise{$37.31\%$ / $53.35\%$} \\
% \vspace{1mm}
\hdashline
% \vspace{1mm}
\revise{+ Reward Design [$\beta_{0/1}=(0.0,0.2)$]} & \revise{$R_{train}$} &
\revise{$28.30\%$ / $29.34\%$} & \revise{$5.68$ / $10.12$} &
\revise{$44.60\%$} & \revise{$37.39\%$ / $52.10\%$} \\
\revise{+ Reward Design [$\beta_{0/1}=(0.0,0.5)$]} & \revise{$R_{train}$} &
\revise{$29.08\%$ / $29.95\%$} & \revise{$5.88$ / $10.53$} &
\revise{$45.16\%$} & \revise{$37.79\%$ / $52.84\%$} \\
\revise{+ Reward Design [$\beta_{0/1}=(0.0,0.8)$]} & \revise{$R_{train}$} &
\revise{$29.60\%$ / $29.34\%$} & \revise{$5.89$ / $10.23$} &
\revise{$45.39\%$} & \revise{$38.68\%$ / $52.38\%$} \\
\hline
\end{tabular}}
\vspace{-4mm}
\end{table*}

% \begin{table*}[t]
% \centering
% \caption{Ablation on the difficulty-aware weights. ``$S.$'' represents results for short questions, while ``$L.$'' denotes results for long questions.}
% \label{tab:ablation-beta}
% \resizebox{\linewidth}{!}{
% \setlength{\tabcolsep}{0.5mm}
% \begin{tabular}{llcccc}
% \hline
% \multirow{2}{*}{\textbf{Model}}
% & \multirow{2}{*}{\textbf{Training Data}}
% & \multicolumn{2}{c}{\textbf{\dataset ($S. / L.$)}}
% & \multicolumn{2}{c}{\textbf{\datasetplus}} \\
% & & Weighted Acc. & Weighted Map Score 
% & Weighted Acc. & Weighted Acc. (Count / TorF) \\
% \midrule\midrule
% Qwen2.5-VL-7B-Instruct & -- &
% $13.28\%$ / $7.12\%$ & $4.01$ / $5.74$ &
% $44.21\%$ & $37.39\%$ / $51.32\%$ \\

% + Reward Design [$\beta_{0/1}=(0.0,0.2)$] & $R_{train}$ &
% $28.30\%$ / $29.34\%$ & $5.68$ / $10.12$ &
% $44.60\%$ & $37.39\%$ / $52.10\%$ \\

% + Reward Design [$\beta_{0/1}=(0.0,0.5)$] & $R_{train}$ &
% $29.08\%$ / $29.95\%$ & $5.88$ / $10.53$ &
% $45.16\%$ & $37.79\%$ / $52.84\%$ \\

% + Reward Design [$\beta_{0/1}=(0.0,0.8)$] & $R_{train}$ &
% $29.60\%$ / $29.34\%$ & $5.89$ / $10.23$ &
% $45.39\%$ & $38.68\%$ / $52.38\%$ \\
% \hline
% \end{tabular}}
% \end{table*}

\begin{table*}[!h]
\centering
\vspace{-2mm}
\caption{Evaluation of \method across model scales. ``$S.$'' represents results for short questions, while ``$L.$'' denotes results for long questions.}
\label{tab:ablation-modelscale}
\resizebox{\linewidth}{!}{
\setlength{\tabcolsep}{0.5mm}
\begin{tabular}{llcccc}
\hline
\multirow{2}{*}{\textbf{Model}} 
& \multirow{2}{*}{\textbf{Training Data}} 
& \multicolumn{2}{c}{\textbf{\dataset ($S. / L.$)}} 
& \multicolumn{2}{c}{\textbf{\datasetplus}}\\ 
& &  Weighted Acc. & Weighted Map Score 
& Weighted Acc. & Weighted Acc. (Count / TorF)  \\ 
\midrule\midrule
Qwen2.5-VL-3B-Instruct  & - & 
$8.68\%$ / $7.99\%$  & $2.75$ / $3.70$ & 
$37.61\%$ & $22.68\%$ / $53.16\%$  \\
+ RL (baseline) & $R_{train}$ & 
$11.46\%$ / $10.50\%$ & $3.81$ / $6.09$ & 
$38.29\%$ & $22.06\%$ / $55.19\%$ \\
+ \method & $RPlus_{train}$ + $R_{train}$ &
$\mathbf{19.36\%}$ / $\mathbf{15.89\%}$ & $\mathbf{4.79}$ / $\mathbf{7.53}$ & 
$\mathbf{65.91\%}$ & $\mathbf{63.58\%}$ / $\mathbf{68.34\%}$ \\
\hline
\end{tabular}}
\vspace{-4mm}
\end{table*}

\subsection{Diagnostic Experiments}
\label{sec:diagnostic-experiments}

We provide extensive ablation studies using Qwen2.5-VL-7B-Instruct, and further verify the effectiveness of \method in addressing sparse-reward issues, across different model scales (\textit{e.g.}, Qwen2.5-VL-3B-Instruct) and \revise{different model architectures (\textit{e.g.}, Kimi-VL-A3B-Instruct).}
% \revise{We further conduct an ablation study on the hyperparameter $\alpha$ and the difficulty-aware weighting scheme.}

\textbf{Ablation on Reward Design and Multi-Stage Design of \method.}
We ablate the reward design and multi-stage design of \method under two training data settings. For the configuration using only \dataset training data, where the multi-stage design cannot be applied, we ablate the reward design alone. As shown in Table~\ref{tab:ablation-rewardmap}, enabling either component individually yields performance gains on both \dataset and \datasetplus, while combining them achieves the best results, confirming the effectiveness and complementarity of the two components.

\textbf{Ablation on \datasetplus.}
Next, we assess the impact of \datasetplus under two training strategies (\textit{e.g.}, SFT and RL). As shown in Table~\ref{tab:ablation-datasetplus}, conducting cold-start training with \datasetplus consistently improves performance on both \dataset and \datasetplus, regardless of the chosen strategy.

\textbf{Ablation on Granularity of Multi-Stage Design.}
We further conduct an ablation study on the granularity of the multi-stage design by comparing a coarse-grained variant with our proposed \method. As shown in Table~\ref{tab:ablation-granularity}, switching to the coarse strategy leads to performance degradation while still outperforming the baseline, thereby reinforcing the effectiveness of the multi-stage design.

\revise{\textbf{Ablation on the Hyperparameter $\alpha$.}
Based on the full \method pipeline, we varied only $\alpha$. The results are reported in Table~\ref{tab:ablation-alpha}. We observe a slight performance drop when $\alpha$ is small, while the performance becomes similar once $\alpha$ enters a higher range. These results suggest that the detailed reward is indeed beneficial (as $\alpha$ determines its contribution to the total reward), and they also indicate that $\alpha$ is not sensitive within a reasonable interval.}

\begin{wrapfigure}{r}{0.5\linewidth}
\vspace{-5mm}
\centering
\includegraphics[width=0.5\textwidth]{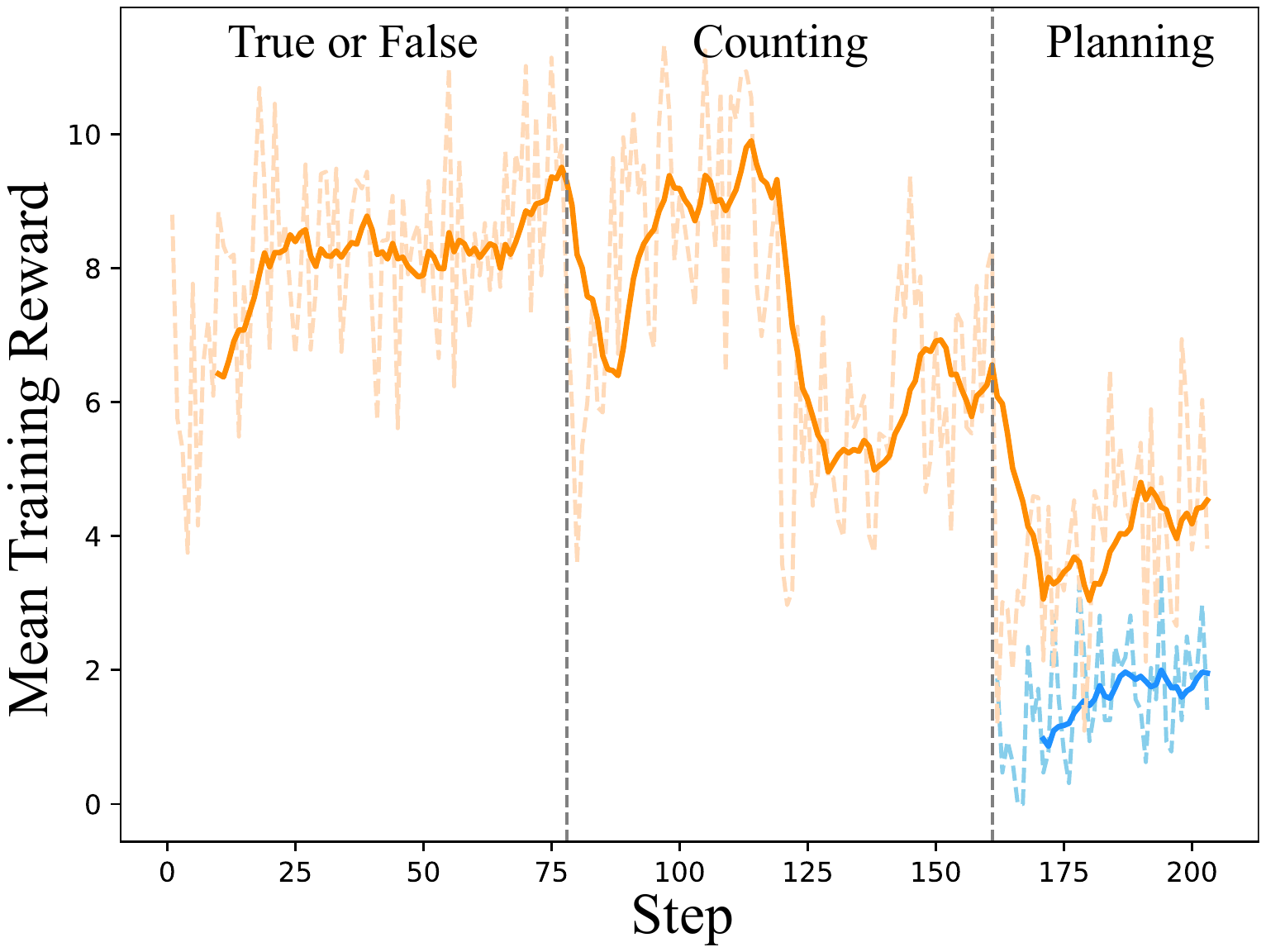}
\vspace{-5mm}
\caption{Comparison of training rewards between baseline RL and \method. The yellow curve denotes the reward trajectory of \method, while the blue curve corresponds to the baseline RL trained solely on \dataset.}
\vspace{-5mm}
\label{fig:ablation-sparse-reward}
\end{wrapfigure}

\revise{\textbf{Ablation on the Difficulty-Aware Weighting Scheme.}
Using the \dataset training set, we further conduct two ablations: one varying only the relative magnitudes of $\gamma_{e/m/h}$ and another varying only those of $\beta_{0/1}$. As shown in Table~\ref{tab:ablation-diff-weights}, performance decreases slightly when the weights become nearly uniform, but remains stable once clear difficulty distinctions are introduced. These results support the effectiveness of our difficulty-aware weighting—where $\gamma_{e/m/h}$ and $\beta_{0/1}$ modulate map and question difficulty, respectively—and indicate that both weights are insensitive within a reasonable range.}

\textbf{Effectiveness of Tackling Sparse Reward.}
We validate the effectiveness of \method to address sparse rewards. As shown in Figure~\ref{fig:ablation-sparse-reward}, we compare reward trajectories of \method with a baseline RL trained on \dataset, aligned at the planning stage. The results show that \method alleviates reward sparsity, further confirming its effectiveness.

\textbf{Effectiveness of \method across Model Scales.}
We evaluate the effectiveness of \method across different model scales. Due to training cost constraints, we adopt Qwen2.5-VL-3B-Instruct as the base model and compare the baseline RL with \method. As shown in Table~\ref{tab:ablation-modelscale}, \method achieves the promising results, demonstrating its robustness and effectiveness.

\begin{table*}[!h]
\centering
\vspace{-3mm}
\caption{\revise{Results on Kimi-VL (Kimi-VL-A3B-Instruct). ``$S.$'' represents results for short questions, while ``$L.$'' denotes results for long questions.}}
\label{tab:kimi-results}
\resizebox{\linewidth}{!}{
\setlength{\tabcolsep}{0.5mm}
\begin{tabular}{llcccc}
\hline
\multirow{2}{*}{\textbf{Model}}
& \multirow{2}{*}{\textbf{Training Data}}
& \multicolumn{2}{c}{\textbf{\dataset ($S. / L.$)}}
& \multicolumn{2}{c}{\textbf{\datasetplus}} \\
& & Weighted Acc. & Weighted Map Score
& Weighted Acc. & Weighted Acc. (Count / TorF) \\
\midrule\midrule
Kimi-VL-A3B-Instruct & - &
$12.76\%$ / $12.33\%$ & $3.30$ / $5.37$ &
$32.55\%$ & $14.75\%$ / $51.08\%$ \\
\revise{+ \method} & \revise{$R_{train}$} &
\revise{$18.58\%$ / $17.36\%$} & \revise{$4.70$ / $7.69$} &
\revise{$35.92\%$} & \revise{$15.20\%$ / $57.50\%$} \\
% + \method & $RPlus_{train}$ + $R_{train}$ &
% --- / --- & $x$ / $x$ &
% $\%$ & $\%$ / $\%$ \\
\hline
\end{tabular}}
\vspace{-3mm}
\end{table*}

\revise{\textbf{Generalization of \method across Model Architectures.}
We further trained Kimi-VL (Kimi-VL-A3B-Instruct) with the \method pipeline using the training set from \dataset. As shown in Table~\ref{tab:kimi-results}, the model achieves substantial performance gains, demonstrating that our method generalizes beyond the Qwen2.5-VL series.}

\section{Conclusion}
\label{sec:conclusion}

In this work, we address the challenge of applying reinforcement learning to fine-grained visual reasoning, where sparse rewards and long reasoning horizons often hinder effective optimization. 
Building upon the \dataset benchmark, we introduce \datasetplus, an extended dataset that organizes tasks along a difficulty continuum, providing dense supervision to facilitate cold-start training. 
Furthermore, we propose \method, a multi-stage reinforcement learning framework that combines curriculum-style task scheduling with difficulty-aware reward design.
Our experiments demonstrate that each of these components contributes to stable and effective training, and that their integration yields the strongest improvements. 
\method not only advances performance on \dataset and \datasetplus but also enhances robustness across broader visual reasoning benchmarks, indicating improved perceptual and reasoning capabilities of multimodal models.
% Overall, this study highlights the importance of designing both curriculum-structured data and reward-aware RL paradigms for training MLLMs in structured visual reasoning tasks. We believe \method provides a principled direction for overcoming sparse reward challenges and can serve as a foundation for future research on curriculum-based reinforcement learning in multimodal AI.

\clearpage

\section*{Acknowledgement}
\label{sec:acknowledgement}

This paper is supported by Young Scientists Fund of the National Natural Science Foundation of China (NSFC) (No. 62506305), Zhejiang Leading Innovative and Entrepreneur Team Introduction Program (No. 2024R01007), Key Research and Development Program of Zhejiang Province (No. 2025C01026), Scientific Research Project of Westlake University (No. WU2025WF003), Chinese Association for Artificial Intelligence (CAAI) \& Ant Group Research Fund - AGI Track (No. 2025CAAI-ANT-13). It is also supported by the research funds of the National Talent Program and Hangzhou Municipal Talent Program.

\section*{Ethics Statement}
\label{sec:ethics-statement}

This work does not involve human subjects or sensitive personal data. \datasetplus is constructed from publicly available transit maps with automatically generated question–answer pairs, ensuring no privacy or security concerns. The datasets are released exclusively for academic research under the Apache License 2.0 on HuggingFace, and all reported results are fully reproducible with the released code and configurations.

\section*{Reproducibility Statement}
\label{sec:reproducibility-statement}

To ensure reproducibility, we provide the evaluation setup details in Section~\ref{sec:experimental-setup} and Appendix~\ref{apx:evaluation-details}, including hardware and implementation, which facilitates rapid replication. 
% We also release \datasetplus\footnote{\url{https://huggingface.co/datasets/ReasonMap-Plus/ReasonMap-Plus}} for reference. All links are anonymized and included as supplementary materials during the review phase, and will be replaced with permanent public links upon acceptance.

\clearpage

% \bibliography{iclr2026_conference}
\bibliographystyle{iclr2026_conference}

\clearpage
\newpage
\appendix
\setcounter{figure}{0}
\setcounter{table}{0}
\renewcommand{\thefigure}{A\arabic{figure}}
\renewcommand{\thetable}{A\arabic{table}}
\section*{Appendix}
\label{apx:apx}

In Appendix~\ref{apx:dataset_details}, we provide a statistical overview of \datasetplus, question templates, and fine-grained annotation. Appendix~\ref{apx:method-detail} outlines the computation pipeline of the detail reward. Appendix~\ref{apx:evaluation-details} describes evaluation settings, including difficulty-aware weighting, benchmark datasets, and result sources in Table~\ref{tab:main-multi-benchmark}. In Appendix~\ref{apx:supplementary-results}, we provide more cases to support our claims in Section~\ref{sec:qualitative-results}. We discuss future work in Appendix~\ref{sec:future-work}. Finally, Appendix~\ref{apx:llm-usage} presents the statement on LLM usage.

\vspace{-0.2cm}
\startcontents[appendices]
\printcontents[appendices]{l}{1}{\setcounter{tocdepth}{3}}

\section{Dataset Construction Details}
\label{apx:dataset_details}

\subsection{Statistical Overview}
\label{apx:statistical-overview}

\datasetplus mirrors \dataset in image composition, comprising high-resolution transit maps from $30$ cities. It contains $4,018$ questions across five categories (\textit{e.g.}, Local Counting 1 - $1,050$; Local Counting 2 - $970$; Global Counting - $30$; True or False 1 - $1,039$; True or False 2 - 929), with a difficulty distribution of $1,259$ easy, $1,342$ middle, and $1,417$ hard. To preserve difficulty balance and diversity, we adopt the same split as \dataset, using questions from $11$ cities for the test set ($1,448$) and the remainder for the training set ($2,570$).

\subsection{Question Template}
\label{apx:question_template}

We present the question templates of \datasetplus as follows.

\vspace{0.5mm}
\begin{tcolorbox}[title=Local Counting 1,colback=gray!5!white,colframe=gray!60!black,fonttitle=\bfseries]
Please solve the multiple choice problem and put your answer (one of ABCD) in one ``\\boxed{}". According to the subway map, how many intermediate stops are there between {stop 1} and {stop 2} (except for this two stops)?
A) {x}
B) {x}
C) {x}
D) {x}
\end{tcolorbox}

\vspace{0.5mm}
\begin{tcolorbox}[title=Local Counting 2,colback=gray!5!white,colframe=gray!60!black,fonttitle=\bfseries]
Please solve the problem and put your answer in one ``\\boxed{}". According to the subway map, how many lines pass through {stop 1}?
\end{tcolorbox}

\vspace{0.5mm}
\begin{tcolorbox}[title=Global Counting,colback=gray!5!white,colframe=gray!60!black,fonttitle=\bfseries]
Please solve the problem and put your answer in one ``\\boxed{}". According to the subway map, how many subway (metro) lines are there in total?
\end{tcolorbox}

\vspace{0.5mm}
\begin{tcolorbox}[title=True or False 1,colback=gray!5!white,colframe=gray!60!black,fonttitle=\bfseries]
Please solve the problem and put your answer (only answer yes or no) in one ``\\boxed{}". According to the subway map, is it true that {stop 1} is the same line as {stop 2}?
\end{tcolorbox}

\vspace{0.5mm}
\begin{tcolorbox}[title=True or False 2,colback=gray!5!white,colframe=gray!60!black,fonttitle=\bfseries]
Please solve the problem and put your answer (only answer yes or no) in one ``\\boxed{}". According to the subway map, is it true that {stop 1} is on the {line x}?
\end{tcolorbox}

\subsection{Fine-Grained Annotation}
\label{apx:fine-grained-annotation}

\revise{We provide more details of the process-level annotation in this section. The construction involves two steps. First, we annotate the route--stop structure of each transit map, including the ordered stops for every line, interchange stations, and branching points. These annotations, as listed in the Meta Data, are recorded in a JSON file. An excerpt for Singapore is shown below:}
\vspace{0.5mm}
\begin{tcolorbox}[title=Meta Data,colback=blue!5!white,colframe=blue!60!black,fonttitle=\bfseries]
``Circle Line": [``HarbourFront (Transfer)", ``Telok Blangah", ``Labrador Park", ...], \\
``Downtown Line": [...], \\
... \\
\end{tcolorbox}
\revise{Based on the Meta Data, we generate process-level annotations with task-specific Python scripts. For example, in the Local Counting~1 category of \datasetplus, where the question asks for the number of intermediate stops between two stations, the script identifies the two queried stops, extracts all intermediate stops between them, computes the count as the final answer, and records the intermediate stops as the process-level annotation. Other task types, such as True/False or shortest-path queries, follow a similar procedure.}
\vspace{0.5mm}
\begin{tcolorbox}[title=Example 1 (Local Counting 1),colback=blue!5!white,colframe=blue!60!black,fonttitle=\bfseries]
  ``country": ``singapore", \\
  ``city": ``singapore", \\
  ``station\_1": ``Clarke Quay", \\
  ``station\_2": ``Farrer Park", \\
  ``figure": ``./maps/singapore/singapore.png", \\
  ``question": ``According to the map, how many intermediate stops are there \\
               between Clarke Quay and Farrer Park (excluding the endpoints)?", \\
  ``answer": ``2", \\
  ``fine-grained answer": [``Dhoby Ghaut", ``Little India"], \\
  ``type": ``counting\_1", \\
  ``difficulty\_city": ``hard", \\
  ``city\_line\_count": ``6", \\
  ``city\_transfer\_count": ``53", \\ 
  ``json": ``./stations/singapore/singapore.json" \\
\end{tcolorbox}
\vspace{0.5mm}
\begin{tcolorbox}[title=Example 2 (True or False 1),colback=blue!5!white,colframe=blue!60!black,fonttitle=\bfseries]
{
  ``country": ``singapore", \\
  ``city": ``singapore", \\
  ``station\_1": ``Somerset", \\
  ``station\_2": ``Bras Basah", \\
  ``figure": ``./maps/singapore/singapore.png", \\
  ``question": ``According to the map, is it true that Somerset is \\
               on the same line as Bras Basah?", \\
  ``answer": ``No", \\
  ``fine-grained answer": {
      ``Somerset": ``North South Line",
      ``Bras Basah": ``Circle Line"
  }, \\
  ``type": ``torf\_1", \\
  ``difficulty\_city": ``hard", \\
  ``city\_line\_count": ``6", \\
  ``city\_transfer\_count": ``53", \\
  ``json": ``./stations/singapore/singapore.json" \\
}
\end{tcolorbox}

\section{\method Details}
\label{apx:method-detail}

\subsection{Computation Pipeline of Detail Reward}
\label{apx:detail-reward}

We present the complete computation pipeline of Detail Reward in Algorithm~\ref{alg:detail-reward}.

\begin{algorithm}[ht]
\caption{Detailed Reward for Planning Questions in \dataset}
\label{alg:detail-reward}

Initialize \texttt{score} $\leftarrow 0$\;

\If{\texttt{route\_data} is empty \textbf{or} format is wrong}{
  \Return \texttt{score}\;
}

\If{departure stop of first segment $=$ stop$\_1$ \textbf{or} arrival stop of last segment $=$ stop$\_2$}{
  \texttt{score} $\leftarrow$ \texttt{score} $+ 2$\;
}

\ForEach{segment $s_i$ in predicted route}{
  \If{\texttt{current\_transfer\_times} $>$ \texttt{question\_transfer\_count}}{
    \texttt{score} $\leftarrow$ \texttt{score} $- 5$\;
  }

  \If{\texttt{current\_transfer\_times} $= 0$ \textbf{and} \texttt{is\_correct}(route name)}{
    \texttt{score} $\leftarrow$ \texttt{score} $+ 4$\;
  }

  \If{departure stop $\in$ \texttt{stations} \textbf{and} arrival stop $\in$ \texttt{stations}}{
    \If{segment $s_i$ is not the last}{
      \If{ arrival stop  $=$ departure stop of next segment}{
        \texttt{score} $\leftarrow$ \texttt{score} $+ 1$\;
      }
    }
  }
}

\texttt{score} $\leftarrow \min(\texttt{score},\,10)$\;
\Return \texttt{score}\;
\end{algorithm}

\section{Evaluation Details}
\label{apx:evaluation-details}

\subsection{Details of Difficulty-Aware Weighting}
\label{apx:weighting-details}

For \dataset, we follow the weighting scheme described in the original paper (see Appendix B.3 therein). For \datasetplus, weights are assigned solely based on map difficulty, with values of $1.0$, $1.5$, and $2.0$ for easy, medium, and hard maps, respectively.

\subsection{Details of Evaluation Datasets}
\label{apx:evaluation-datasets-details}

We provide a brief description of the six evaluation benchmarks used in our paper:

\begin{enumerate}
    \item \textbf{SEED-Bench-2-Plus (Map)} \citep{li2024seednbench2} describes three categories (Charts, Maps, Webs) with human-verified multiple-choice items, from which we use the Map slice.
    \item \textbf{SpatialEval} \citep{wang2024SpatialEval} targets spatial intelligence across relationships, position, counting, and navigation. 
    \item \textbf{$V^*$Bench} \citep{wu2024vstar} evaluates fine-grained attribute recognition and spatial relationships on high-resolution images.
    \item \textbf{HRBench} \citep{wang2025HRBench} evaluates MLLMs on 4K/8K high-resolution images and introduces a training-free enhancement baseline.
    \item \textbf{ChartQA} \citep{masry2022chartqa} benchmarks QA over charts with visual and logical reasoning.
    \item \textbf{MMStar} \citep{chen2024MMStar} offers 1,500 human-curated, vision-indispensable samples covering 6 core capabilities and 18 axes.
\end{enumerate}
%
% $\mathbf{V^{*}}$ and \textbf{HRBench} stress fine-grained, region-sensitive perception and compositional reasoning.
% %
% \textbf{ChartQA} benchmarks QA over charts with visual and logical reasoning, while \textbf{MMStar} offers 1,500 human-curated, vision-indispensable samples covering 6 core capabilities.

\subsection{Result Source Summary}
\label{apx:result-source}

We present sources of results in Table~\ref{tab:main-multi-benchmark}: $\dagger, \ddagger, \$, *, \S$ correspond to the Seed1.5-VL technical report~\citep{guo2025seed1}, the Qwen2.5-VL technical report~\citep{bai2025qwen25}, the InternVL3.5 technical report~\citep{wang2025internvl35}, the Kimi-VL technical report~\cite{team2025kimi}, and the official Hugging Face repository of Qwen2.5-VL-32B-Instruct\footnote{\url{https://huggingface.co/Qwen/Qwen2.5-VL-32B-Instruct}}, respectively.

\revise{\section{Supplementary Results}
\label{apx:supplementary-results}

\subsection{Comparison Cases}
\label{apx:comparison-cases}

We provide more comparison cases to support deeper analysis in Section~\ref{sec:qualitative-results}. In the following case, the prediction from Qwen2.5-VL-7B-Instruct exhibits a visual confusion error at \textit{$<$wrong1$>$} and a hallucination at \textit{$<$wrong2$>$}, misidentifying three intermediate stops but selecting option~A (6). In contrast, the RewardMap-trained model correctly identifies a single intermediate stop and selects the correct option~C.

\begin{tcolorbox}[colback=gray!2,colframe=gray!40,rounded corners]
\textbf{Question:} 
\textit{How many intermediate stops are there between A Station and B Station?}
(\texttt{<budapest.png>} + Format Requirement + \{A: 6, B: 8, C: 1, D: 9\})
\end{tcolorbox}

\begin{tcolorbox}[colback=white,colframe=red!50,rounded corners]
\textbf{Answer (Qwen2.5-VL-7B-Instruct, Wrong).} \\
To determine the number of intermediate stops between A Station and B Station, we need to look at the yellow line on the map, which represents Metro Line~1.

1. Identify the stations on the yellow line between A and B.  
2. Count the number of stations, excluding A and B.

From the map:  
-- A \\
-- Széchenyi Thermal Bath \\
-- Bajza Street \\
-- B  

There are 3 intermediate stops between A and B. \textit{$<$wrong1$>$}  

Therefore, the correct answer is:  
\(\boxed{A}\) \textit{$<$wrong2$>$}
\end{tcolorbox}

\begin{tcolorbox}[colback=white,colframe=green!45!black,rounded corners]
\textbf{Answer (Qwen2.5-VL-7B-Instruct + RewardMap, Correct).} \\
To solve this problem, we need to count the intermediate stops between A Station and B Station, excluding these two stops.

Starting from A:  
-- The next stop is Széchenyi Thermal Bath.

Thus, the only intermediate stop between A and B is Széchenyi Thermal Bath.

Therefore, the correct answer is:  
\(\boxed{C}\)
\end{tcolorbox}}
\revise{\section{Future Work}
\label{sec:future-work}

Our framework is not limited to transit maps and can be extended to other structured visual domains such as charts and diagrams. Our results support this direction: ChartQA~\citep{masry2022chartqa} improves accuracy from $86.12\%\rightarrow87.24\%$, Charxiv~\citep{wang2024charxiv} from $66.41\%\rightarrow68.90\%$, and InfoVQA~\citep{mathew2022infographicvqa} from $82.22\%\rightarrow82.71\%$. These gains suggest that the detail rewards and multi-stage RL scheme generalize to structurally similar tasks. Future work will explore domain-agnostic rewards and broader applications across diverse structured visual reasoning settings.

Efficient methods can further improve the fine-tuned model by our framework~\citep{feng2024oracle,bai2025ressvd,zhu2025obs,shao2025holitom,shao2025tokens,tao2025dycoke,tao2025omnizip,tao2025plug}. The idea of detail reward can be extended to other fields that require reasoning capability~\citep{song8,song10,wang2025pointlora,wang2025forging,ai2025lightweight,wu2025niagara,Li_2025_CVPR,wang2023lidar2map,jin2025mergemix,xiao2025occlusion,li2025coser,feng2026dvoting,wang2026free}. Leveraging reinforcement learning for safety alignment~\citep{zhang2025poison,li2026sponge,li2025vid} remains a promising direction for further exploration. 

}
% \section{Further Statement}
% \label{apx:license-content-info}

\section{Large Language Model Usage Statement}
\label{apx:llm-usage}

Large Language Models (LLMs) were used only for surface-level editing of the manuscript (e.g., polishing grammar and style, rephrasing for clarity, and making minor \LaTeX{} adjustments). They were not involved in generating ideas, methods, algorithms, code, experiments, figures, tables, or citations. All research design, implementation, data processing, and analysis were performed by the authors. LLM use was limited to de-identified text snippets, with no proprietary data or unpublished results shared, and all outputs were manually reviewed and revised. This limited assistance does not affect reproducibility, as every reported result is fully reproducible from the released code and configurations.

\end{document}